\documentclass{article}
\usepackage[utf8]{inputenc}

\usepackage{authblk}
\usepackage[hidelinks]{hyperref}
    
\title{Transgressing the boundaries: towards a rigorous \\understanding of deep learning and its (non-)robustness}
 \author[1]{Carsten Hartmann}
 \author[2,3]{Lorenz Richter}
\date{\today}

 \affil[1]{Institute of Mathematics, BTU Cottbus-Senftenberg, 03046 Cottbus, Germany, \href{mailto:carsten.hartmann@b-tu.de}{carsten.hartmann@b-tu.de}}
 \affil[2]{Zuse Institute Berlin, 14195 Berlin, Germany, \href{mailto:richter@zib.de}{richter@zib.de}}
 \affil[3]{dida Datenschmiede GmbH, 10827 Berlin, Germany}

\usepackage{mathrsfs, amssymb}
\usepackage{amsthm}
\usepackage[intlimits]{empheq}
\usepackage{bbm}     \usepackage{amsmath}
\usepackage{xcolor}
\usepackage[scientific-notation=true]{siunitx}
\usepackage{float}
\usepackage{dsfont}	
\usepackage[authoryear]{natbib}
\usepackage[nottoc,numbib]{tocbibind}
\usepackage{bbm}
\usepackage{subcaption}
\usepackage[inner=2cm,outer=2cm,top=2cm,bottom=2cm]{geometry}
\usepackage[ruled,vlined]{algorithm2e}
\usepackage{enumerate}   
\usepackage{cleveref}

\DeclareMathOperator*{\argmin}{arg\,min}
\let\P\relax\newcommand{\P}{\mathbb{P}}

\DeclareMathOperator{\E}{\mathbb{E}}
\DeclareMathOperator*{\R}{\mathbb{R}}

\theoremstyle{plain}
\newtheorem{theorem}{Theorem}[section]

\theoremstyle{definition}
\newtheorem{definition}[theorem]{Definition}
\newtheorem{example}[theorem]{Example}

\theoremstyle{remark}

\usepackage[T1]{fontenc}
\usepackage{lipsum}
\usepackage{tocloft}
\begin{document}
    
\maketitle

\begin{abstract}
 The recent advances in machine learning in various fields of applications can be largely attributed to the rise of deep learning (DL) methods and architectures. Despite being a key technology behind autonomous cars, image processing, speech recognition, etc., a notorious problem remains the lack of theoretical understanding of DL and related interpretability and (adversarial) robustness issues. Understanding the specifics of DL, as compared to, say, other forms of nonlinear regression methods or statistical learning, is interesting from a mathematical perspective, but at the same time it is of crucial importance in practice: treating neural networks as mere black boxes might be sufficient in certain cases, but many applications require  waterproof performance guarantees and a deeper understanding of what could go wrong and why it could go wrong. 
 It is probably fair to say that, despite being mathematically well founded as a method to approximate complicated functions, DL is mostly still more like modern alchemy that is firmly in the hands of engineers and computer scientists. Nevertheless, it is evident that certain specifics of DL that could explain its success in applications demands systematic mathematical approaches. 
In this work, we review robustness issues of DL and particularly bridge concerns and attempts from approximation theory to statistical learning theory. Further, we review Bayesian Deep Learning as a means for uncertainty quantification and rigorous explainability.
\end{abstract}

\section{Introduction}

According to \citet[p.~2]{wheeler2016}, machine learning is a ``marriage of statistics and computer science that began in artificial intelligence''. While statistics deals with the question of what can be inferred from data given an appropriate statistical model, computer science is concerned with the design of algorithms to solve a given computational problem that would be intractable without the help of a computer.         

Artificial intelligence and, specifically, machine learning have undergone substantial developments in recent years that have led to a huge variety of successful applications, most of which would not have been possible with alternative approaches. In particular, advances in deep learning (i.e. machine learning relying on deep neural networks) have revolutionized many fields, leading, for instance, to impressive achievements in computer vision (e.g. image classification, image segmentation, image generation), natural language processing (semantic text understanding, text categorization and text creation, automatic question answering) and reinforcement learning (agents and games, high-dimensional optimization problems); cf.~\citet{sarker2021review} and the references therein.

Moreover, deep learning is nowadays increasingly applied in multiple scientific branches as an acceptable tool for conducting inference from simulated or collected data. For example, in the medical field, the development of drugs \citep{ma2015deep} or the analysis of tomography \citep{bubba2019learning} are enhanced with deep learning. In molecular simulations, ground-state properties of organic molecules are predicted \citep{faber2017prediction}, equilibrium energies of molecular systems are learnt \citep{noe2019boltzmann} or multi-electron Schr\"odinger equations are solved \citep{hermann2020deep}. Speaking of which, the numerical treatment of high-dimensional partial differential equations with neural networks has undergone vast improvements \citep{weinan2017deep, nusken2021solving}, allowing for applications in almost all sciences. In biology, cell segmentation and classification have been studied with certain convolutional neural networks \citep{ronneberger2015u}, in signal processing speech separation is approached with temporal versions of these \citep{richter2020speech}, and in finance relevant stock pricing models are solved with deep learning \citep{germain2021neural}. In remote sensing, temporal recurrent neural networks are for instance used for crop classification \citep{russwurm2018multi} and image segmentation promises automatic understanding of the increasing amount of available satellite data \citep{zhu2017deep}. The list of successful deep learning applications is long and there are many more fields in which they have made significant contributions and still promise exciting advances that we shall omit here for the sake of brevity.

It is probably fair to say that, like statistics, deep learning (or machine learning in general) aims at drawing inferences from data. But unlike statistics, it avoids being overly explicit regarding the underlying model assumptions. In statistics, either the model assumptions or the complete model are set prior to making inferences, whereas the neural networks in deep learning are mostly seen as black boxes that are essentially able to `learn' the model. In this sense, deep learning delegates what \citet[\S 72, p.~374]{reichenbach1949prob} called the ``problem of the reference class'' to a computer algorithm, namely, the problem of deciding what model class to use when making a prediction of a particular instance or when assigning a probability to a particular event. While this might be understandable -- or even desirable -- from the user's point of view, it poses risks and might bring dangerous side-effects:
\begin{itemize}
    \item In most of the applied deep learning models, there is a lack of explainability, meaning that even though their inference from data might work well, the mechanisms behind the predictions are not well understood. As the ambition in all sciences it to understand causal relationships rather than pure correlations, this might neither be satisfying nor lead to further deeper understandings in corresponding fields.
    \item Without understanding the details of a model, potential robustness issues might not be realized either. For example, who guarantees that certain deep learning achievements easily translate to slightly shifted data settings and how can we expect neural network training runs to converge consistently?
    \item Finally, often the ambition of a prediction model to generalize to unseen data is stated on an `average' level and we cannot make robust statements on unexpected events, which might imply dangerous consequences in risk-sensitive applications. In general, there is no reliable measure for prediction (un-)certainty, which might lead to blind beliefs in the model output.
\end{itemize}

Even when it comes to the success stories of deep learning, many achievements and properties of the models can simply not be explained theoretically, e.g. why does one of the most naive optimization attempts, stochastic gradient descent, work so well, why do models often generalize well even though they are powerful enough to simply memorize the training data and why can high-dimensional problems be addressed particularly efficiently? Not only is it important from a practical point of view to understand these phenomena theoretically, as a deeper understanding might motivate and drive novel approaches leading to even more successful results in practice, but it is also important for getting a grip on the epistemology of machine learning algorithms. This then might also advance pure `trial and error' strategies for architectural improvements of neural networks that sometimes seem to work mostly due to extensive hyperparameter finetuning and favorable data set selections; cf.~\citep{wang2019convergence}.
\par\bigskip

In this article, we will argue that relying on the tempting black box character of deep learning models can be dangerous and it is important to further develop a deeper mathematical understanding in order to obtain rigorous statements that will make applications more sound and more robust. We will demonstrate that there are still many limitations in the application of artificial intelligence, but mathematical analysis promises prospects that might at least partially overcome these limitations.  
We further argue that, if one accepts that explainable DL must not be understood in the sense of the deductive-nomological model of scientific explanation, Bayesian probability theory can provide a means to explain DL in a precise statistical (abductive) sense. 
In fact, a comprehensive theory should guide us towards coping with the potential drawbacks of neural networks, e.g. the lack of understanding why certain networks architectures work better than others, the risk of overfitting data, i.e. not performing well on unseen data, or the lack of knowledge on the prediction confidences, in particular, leading to overconfident predictions on data far from the training data set.

Even though we insist that understanding deep learning is a holistic endeavor that comprises the theoretical (e.g. approximation) properties of artificial neural networks in combination with the practical numerical algorithms that are used to train them, we refrain from going beyond the mathematical framework and exploring the epistemological implications of this framework. The epistemology of machine learning algorithms is a relatively new and dynamic field of research, and we refer to recent papers by \citet{wheeler2016} and \citet{sterkenburg2021schurz}, and the references given there.

\subsection{Definitions and first principles}

We can narrow down the definition of machine learning to one line by saying that its main intention is to identify functions that map input data $x \in \mathcal{X}$ to output data $y \in \mathcal{Y}$ in some \textit{good} way, where $\mathcal{X}$ and $\mathcal{Y}$ are suitable spaces, often identified with $\R^d$ and $\R$, respectively. In other words, the task is to find a function $f:\mathcal{X} \to \mathcal{Y}$ such that
\begin{equation}
\label{eq: f(x) = y}
    f(x) = y.
\end{equation}

To illustrate, let us provide two stereotypical examples that appear in practice. In a classification task, for instance, $x \in \mathcal{X}$ could represent an image (formalized as a matrix of pixels, or, in a flattened version, as a vector $x \in \R^d$) and $y \in \mathcal{Y} = \{1, \dots, K \}$ could be a class describing the content of the image. In a regression task, on the other hand, one tries to predict real numbers from the input data, e.g. given historical weather data and multiple measurements, one could aim to predict how much it will rain tomorrow and $y\in \mathcal{Y} = \R_{\ge 0}$ would be the amount of rain in milliliters. \par\bigskip

From our simple task definition above, two questions arise immediately:
\begin{enumerate}
    \item How do we design (i.e. find) the function $f$?
    \item How do we measure performance, i.e. how do we quantify deviations from the desired fit in \eqref{eq: f(x) = y}?
\end{enumerate}
Relating to question 1, it is common to rely on parametrized functions $f(x)=f_\theta(x)$, for which a parameter vector $\theta \in \R^p$ specifies the actual function. Artificial neural networks (ANNs) like deep neural networks are examples of such parametrized functions which enjoy specific beneficial properties, for instance in terms of approximation and optimization as we will detail later on. The characterizing feature of (deep) neural networks is that they are built by (multiple) concatenations of nonlinear and affine-linear maps: 

\begin{definition}[Neural network, e.g.~\cite{berner2021modern,higham2019deep}]
\label{def_NN}
We define a \textit{feed-forward neural network} $\Phi_\sigma:\R^d \to \R^m$ with $L$ layers by
\begin{equation}
\Phi_\sigma(x) = A_L \sigma(A_{L-1} \sigma(\cdots  \sigma(A_1 x + b_ 1) \cdots) + b_{L-1}) + b_L,
\end{equation}
with matrices $A_l \in \R^{n_{l} \times n_{l-1}}$, vectors $b_l \in \R^{n_l}, 1 \le l \le L$, and a nonlinear activation function $\sigma: \R \to \R$ that is applied componentwise. Clearly, $n_0=d$ and $n_L=m$, and the collection of matrices $A_l$ and vectors $b_l$, called \emph{weights} and \emph{biases}, comprises the learnable parameters $\theta$.
\end{definition}

In practice, one often chooses $\sigma(x) = \max\{ x, 0\}$ or $\sigma(x) = (1 + e^{-x})^{-1}$, since their (sub)derivatives can be explicitly computed and they enjoy a universal approximation property \citep{barron1993approximation,cybenko1989approximation}.
 
Even though the organization of an ANN in layers is partly inspired by biological neural networks, the analogy between ANNs and the human brain is questionable and often misleading when it comes to understanding the specifics of machine learning algorithms, such as its ability to generalize \citep{wichmann2018humans}, and it will therefore play no role in what follows. We rather regard an ANN as a handy representation of the parametrized function $f_\theta$ that enjoys certain mathematical properties that we will discuss subsequently. (Note that closeness in function space does not necessarily imply closeness in parameter space and vice versa as has been pointed out in \citet[Sec.~2]{elbrachter2019degenerate}.) Clearly, alternative constructions besides the one stated in \Cref{def_NN} are possible and frequently used, depending on the problem at hand. 

\subsection{Probabilistic modelling and mathematical perspectives}  
\label{sec: probabilistic modelling}

Now, for actually tuning the parameter vector $\theta$ in order to identify a good fit as indicated in \eqref{eq: f(x) = y}, the general idea in machine learning is to rely on training data $(x_n, y_n)_{n=1}^N \subset \mathcal{X} \times \mathcal{Y}$. For this, we define a \textit{loss function} $\ell:\mathcal{Y} \times \mathcal{Y} \to \R_{\ge 0}$ that measures how much our predictions, i.e. function outputs $f(x_n)$, deviate from their targets $y_n$. Given the training sample, our algorithm can now aim to minimize the \textit{empirical loss}
\begin{equation}
\label{eq: empirical loss}
    \mathcal{L}_N(f) = \frac{1}{N}\sum_{n=1}^N \ell\left(f(x_n), y_n\right),
\end{equation}
i.e. an empirical average over all data points. Relating to question 2 from above, however, it turns out that it is not constructive to measure approximation quality by how well the function $f$ can fit the available training data, but rather to focus on the ability of $f$ to generalize to yet unseen data. To this end, the perspective of statistical learning theory assumes that the data is distributed according to an (unknown) probability distribution $\P$ on $\mathcal{X} \times \mathcal{Y}$. The training data points $x_n$ and $y_n$ should then be seen as realizations of the random variables $X$ and $Y$, which admit a joint probability distribution, so
\begin{equation}
    (X, Y) \sim \P.
\end{equation}
We further assume that all pairs $(x_n,y_n)$ are distributed identically and independently from one another (i.i.d.). The expectation over all random (data) variables of this loss is then called expected loss, defined as
\begin{equation}
    \mathcal{L}(f) = \E\!\left[\ell\left(f(X), Y\right) \right],
\end{equation}
where the expectation $\E\left[\cdot\right]$ is understood as the average over all possible data points $(X,Y)$. The expected loss measures how well the function $f$ performs on data from $\mathbb{P}$ \textit{on average}, assuming that the data distribution does not change after training. It is the general intention in machine learning to have the expected loss as small as possible.

\begin{example}
To fix ideas, let us consider a toy example in $d=1$. We assume that the true function is given by $f(x) = \sin (2\pi x)$ and that the data $x$ is distributed uniformly on the interval $[0, 2]$. In \Cref{fig: sin joint density} we display the function $f$ along with $N=100$ randomly drawn data points $(x_n, y_n)_{n=1}^N$, where $y_n$ is once given by the deterministic mapping $y_n = f(x_n)$ and once by the stochastic mapping $y_n = f(x_n) + \eta_n$, where $\eta_n \sim \mathcal{N}(0, 0.01)$ indicates noise, by denoting $\mathcal{N}(\mu, \sigma^2)$ a normal (i.e. Gaussian) distribution with mean $\mu$ and variance $\sigma^2$. The stochastic mapping induces the probability measure $\mathbb{P}$, i.e. the joint distribution of the random variables $(X, Y) \in \mathcal{X} \times \mathcal{Y}$, which we plot approximately in the right panel. Note that (even for simple toy problems) $\P$ can usually not be written down analytically.
\begin{figure}[H]
\centering
\includegraphics[width=1\linewidth]{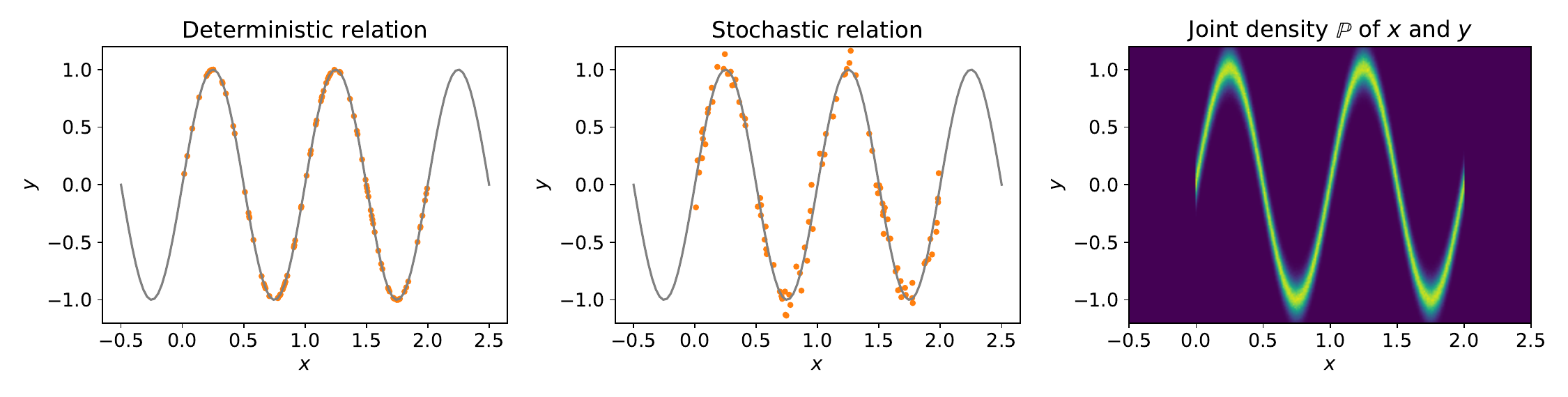}
\caption{We plot a given function $f(x) = \sin(2 \pi x)$ (in gray) along with data points (in orange) given either by a deterministic or stochastic mapping in the first two panels. The right panel shows an approximation of the measure $\P$ for the stochastic case.}
\label{fig: sin joint density}
\end{figure}
\end{example}

For a further analysis, let us give names to three different functions that minimize a given corresponding loss (assuming for simplicity that all minima are attained, even though they may not be unique):
\begin{equation}
    f^B \in \argmin_{f \in \mathcal{M}(\mathcal{X}, \mathcal{Y})} \mathcal{L}(f), \qquad f^* \in \argmin_{f \in \mathcal{F}}\mathcal{L}(f), \qquad \widehat{f}_N \in \argmin_{f\in \mathcal{F}} \mathcal{L}_N(f).
\end{equation}
The first quantity, $f^B$, is the theoretically optimal function among all mathematically reasonable (or: \emph{measurable}) functions (cf. Appendix \ref{sec:bayesopt}), denoted here by the set $\mathcal{M}(\mathcal{X}, \mathcal{Y})$, the second quantity, $f^*$, is the optimal function in a specified function class $\mathcal{F}$ (e.g. the class of neural networks), and the third quantity, $\widehat{f}_N$, is the function that minimizes the empirical error on the training data. 

With regard to the second quantity above, finding a suitable function class $\mathcal{F}$ requires balancing two conflicting goals: on the one hand, the function class should be sufficiently rich to enjoy the \emph{universal approximation property}, i.e. the ability to represent any theoretically optimal function $f^B$ up to a sufficiently small approximation error that is still considered  acceptable.\footnote{What is considered an acceptable approximation error depends on the problem at hand.} On the other hand, the function class should not be overly complex, in order to avoid overfitting which may lead to a function $f$ (e.g. a classifier) that poorly generalizes beyond known data.

Let us make this point more precise, and let us say that we have some training algorithm that has produced a function $f$ on the training data $(x_n, y_n)_{n=1}^N$ (see Appendix \ref{sec:SGD} for details). 

We can decompose the deviation of the function $f$ from the theoretically optimal solution $f^B$ into four different terms that correspond to three different error contributions -- generalization, optimization and approximation error: 
\begin{equation}
\label{eq: error decomposition}
    \mathcal{L}(f) - \mathcal{L}(f^B) = \underbrace{\mathcal{L}(f) - \mathcal{L}_N(f)}_{\text{generalization error}} + \underbrace{\mathcal{L}_N(f) - \mathcal{L}_N({f}^*)}_{\text{optimization error}} + \underbrace{\mathcal{L}_N({f}^*) - \mathcal{L}(f^*)}_{\text{generalization error}} + \underbrace{\mathcal{L}(f^*) - \mathcal{L}(f^B)}_{\text{approximation error}}.
\end{equation}
Specifically, if we set $f=\widehat{f}_N$, the above decomposition reveals what is known as the bias-variance tradeoff, namely, the decomposition of the total error (as measured in terms of the loss) into a contribution that stands for the ability of the function $f^*\in\mathcal{F}$ to best approximate the truth $f^B$ (bias) and a contribution that represents the ability to estimate the approximant $f^*$ from finitely many observations (variance), namely\footnote{Here we loosely understand the word `truth' in the sense of empirical adequacy following the seminal work of van Fraasen \citet[p.~12]{vanFraasen1980scientific}, which means that we consider the function $f^B$ to be empirically adequate, in that there is no other function (e.g. classifier or regression function) that has a higher likelihood relative to all unseen data in the world; see also \citet{hanna1983adequacy}. The term `truth' is typical jargon in the statistical learning literature, and one should not take it as a scientific realist's position.} 
\[
\mathcal{L}(\widehat{f}_N) - \mathcal{L}(f^B) = \underbrace{\mathcal{L}(\widehat{f}_N) - \mathcal{L}(f^*)}_{\text{estimation error (variance)}} + \underbrace{\mathcal{L}(f^*) - \mathcal{L}(f^B).}_{\text{approximation error (bias)}}
\]
We should stress that it is not fully understood yet in which cases overfitting leads to poor generalization and prediction properties of an ANN as there are cases in which models with many (nonzero) parameters that are perfectly fitted to noisy training data may still have good generalization skills; cf.~\citep{bartlett2020overfitting} or Section \ref{sec:generalmemory} below for further explanation. 

A practical challenge of any function approximation and any learning algorithm is to minimize the expected loss by only using a finite amount of training data, but without knowing the underlying data distribution $\mathbb{P}$. In fact, one can show there is no universal learning algorithm that works well for every data distribution (\emph{no free lunch theorem}). Instead, any learning algorithm (e.g. for classification) with robust error bounds must necessarily be accompanied by a priori regularity conditions on the underlying data distribution, e.g.~\citep{berner2021modern,shalev2014understanding,wolpert1996}.\footnote{As a consequence, deep learning does not solve Reichenbach's reference class problem or gives any hint to the solution of the problem of induction, but it is rather an instance in favor of the Duhem-Quine thesis, in that any learning algorithm that generalizes well from seen data must rely on appropriate background knowledge \citep[p.~44]{quine1951}; cf.~\citet{sterkenburg2019putnam}.}     

\par\bigskip

Let us come back to the loss decomposition \eqref{eq: error decomposition}. The three types of errors hint at different perspectives that are important in machine learning from a mathematical point of view:
\begin{enumerate}
	\item Generalization: How can we guarantee generalizing to unseen data while relying only on a finite amount of training data?
    \item Function approximation: Which neural network architectures do we choose in order to gain good approximation qualities (in particular in high-dimensional settings)?
    \item Optimization: How do we optimize a complicated, nonconvex function, like a neural network?
\end{enumerate}

Besides these three, there are more aspects that cannot be read off from equation \eqref{eq: error decomposition}, but turn out to become relevant in particular in certain practical applications. Let us stress the following two:

\begin{enumerate}
\setcounter{enumi}{3}
        \item Numerical stability and robustness: How can we design neural networks and corresponding algorithms that exhibit some numerical stability and are robust to certain perturbations? 
        \item Interpretability and uncertainty quantification: How can we explain the input-output behavior of certain complicated, potentially high-dimensional function approximations and how can we quantify uncertainty in neural network predictions?
\end{enumerate}

In this article, we will argue that perspectives 4 and 5 are often overlooked, but still in particular relevant for a discussion on the limitations and prospects in machine learning. Along these lines, we will see that there are promising novel developments and ideas that advance the aspiration to put deep learning onto more solid grounds in the future. 

The article is organized as follows. In \Cref{sec: ANNs - oddities} we will review some aspects of neural networks, admittedly in a very a non-exhaustive manner, where in particular Sections \ref{sec:generalmemory}--\ref{sec:regularize} will correspond to perspectives 1--3 stated above. \Cref{sec: adversarial attacks} will then demonstrate why (non-)robustness issues in deep learning are particularly relevant for practical applications, as illustrated by adversarial attacks in \Cref{sec:imageattack}. We will argue in \Cref{sec: worst-case scenarios} that successful adversarial attacks on (deep) neural networks require careful thinking about worst-case analyses and uncertainty quantification. \Cref{sec: adversarial attacks} therefore relates to perspectives 4 and 5 from above. Next, \Cref{sec: Bayesian perspective} will introduce the Bayesian perspective as a principled framework to approach some of the robustness issues raised before. After introducing Bayesian neural networks, we will discuss computational approaches in \Cref{sec: BNN in practice} and review further challenges in \Cref{sec: challenges for BNNs}. Finally, in \Cref{sec: conclusion} we will draw a conclusion.

\section{Deep neural networks: oddities and some specifics}
\label{sec: ANNs - oddities}

One of the key questions regarding machine learning with (deep) neural networks is related to their ability to generalize beyond the data used in the training step (cf. perspective 1 in \Cref{sec: probabilistic modelling}). The idea here is that a trained ANN applies the regularities found in the training data (i.e. in past observations) to future or unobserved data, assuming that these regularities are persistent. Without dwelling on technical details, it is natural to understand the training of a neural network from a probabilistic viewpoint, with the trained ANN being a collection of functions, that is characterized by a probability distribution over the parameter space, rather than by a single function. 
This viewpoint is in accordance with how the training works in practice, since training an ANN amounts to minimizing the empirical loss given some training data, as stated in equation \eqref{eq: empirical loss}, and this minimization is commonly done by some form of stochastic gradient descent (SGD) in the high-dimensional \emph{loss landscape}\footnote{The empirical risk $J_N(\theta)=\mathcal{L}_N(f_\theta)$, considered as a function of the parameters $\theta$ is often called the \emph{loss landscape} or \emph{energy landscape}.}, i.e. batches of the full training set are selected randomly during the training iterations (see also \Cref{sec:SGD}). 
As a consequence, the outcome of the training is a random realization of the ANN and one can assign a probability distribution to the trained neural network.

\subsection{Generalization, memorization and benign overfitting}\label{sec:generalmemory}

If we think of the parametrized function that represents a trained neural network as a random variable, it is natural to assign a probability measure $Q(f)$ to every regression function $f$. 
So, let $Q^B=Q(f^B)$ be the target probability distribution (i.e.~the truth), $Q^*=Q(f^*)$ the best approximation, and $\widehat{Q}_N=Q(\widehat{f}_N)$ the distribution associated with the $N$ training points that are assumed to be randomly drawn from $\P$. 
We call $f(t)\in\mathcal{F}$ the function approximation that is obtained after running the parameter fitting until time $t$ (see Sec.~\ref{sec:regularize} and Appendix \ref{sec:SGD} below for further details) -- $f(t)$ therefore models the training for a specified amount of training iterations. Ideally, one would like to see that $Q(f(t))$ resembles either the truth $Q^B$ or its best approximation $Q^*$ as the training proceeds; however, it has been shown that trained networks often memorize (random) training data in that \cite[Thm.~6]{yang2021generalization}  
\[
\lim_{t\to\infty} Q(f(t)) = \widehat{Q}_N\,.
\]
In this case, the training lets the model learn the data which amounts to memorizing facts, without a pronounced ability to generate knowledge. It is interesting to note that this behavior is consistently observed when the network is trained on a completely random relabelling of the true data, in which case one would not expect outstanding generalization capabilities of the trained ANN \citep{zhang2021understanding}. Finally, it so happens that $Q(f(t))$ does not converge to $\widehat{Q}_N$, in which case it diverges and thus gives no information whatsoever about the truth.  \par\bigskip

\begin{figure}[H]
\centering
\includegraphics[width=0.8\linewidth]{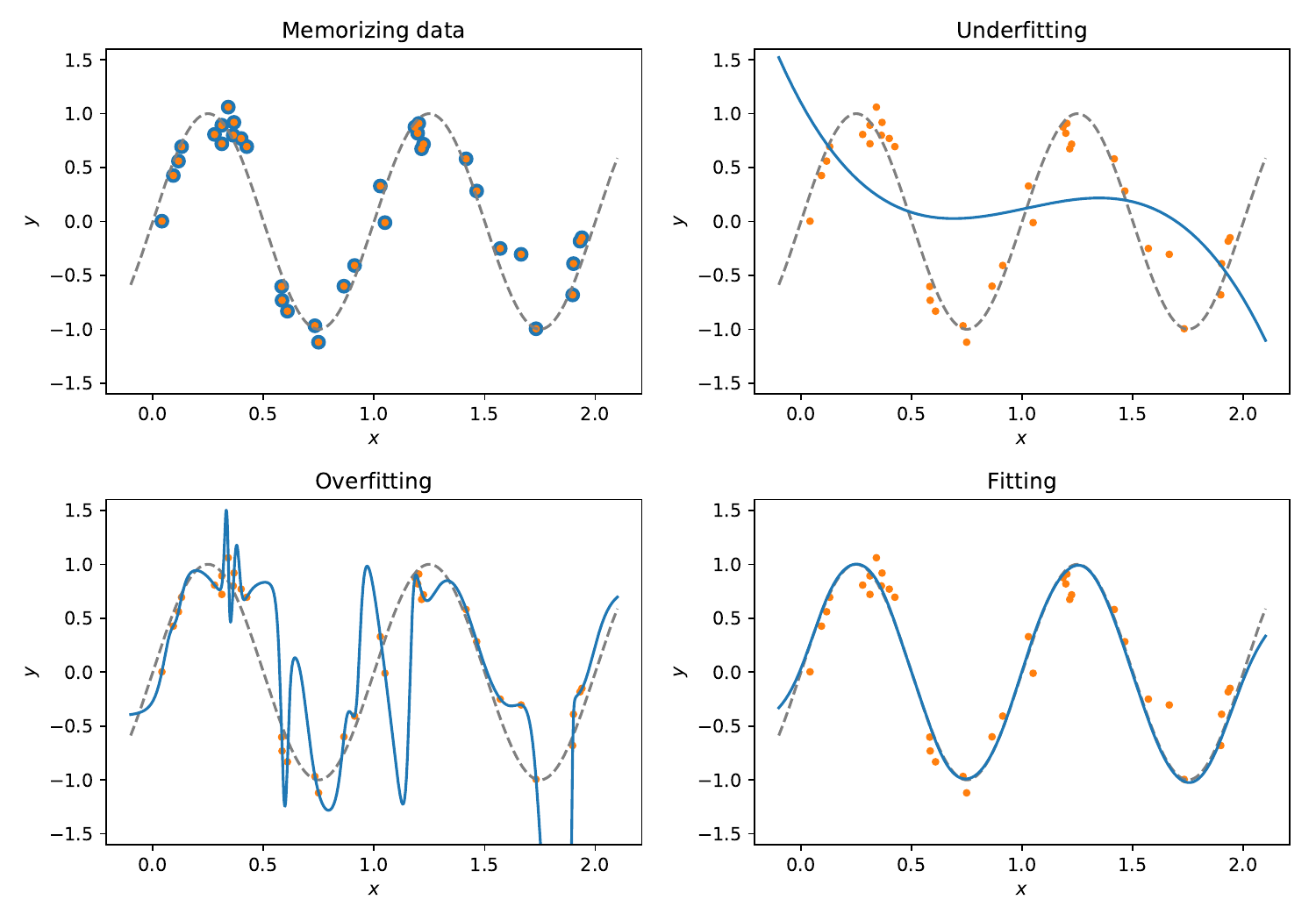}
\caption{Different examples of good and bad fits in the classical regression scheme: While a perfect fit to the training data may either lead to a high-fidelity model on the training data that has no (upper left panel) or very little (lower left panel) predictive power, underfitting leads to a low-fidelity model on the training data (upper right panel). A good fit (lower right panel) is indicated by a compromise between model-fidelity and predictive power.}
\label{fig: underfitting overfitting}
\end{figure}

\begin{figure}[H]
\centering
\includegraphics[width=0.45\linewidth]{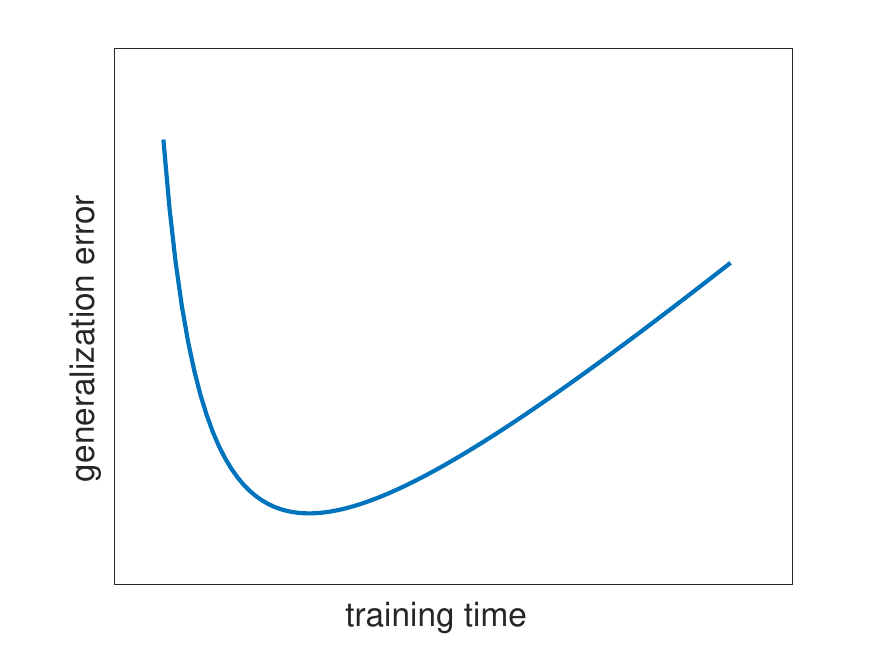}
\caption{Traditional risk curve: schematic sketch of the generalization error of a generic deep neural network for a fixed amount of training data as a function of the training time $t$; see \citep{yang2021generalization} for details.}
\label{fig:training}
\end{figure}

A phenomenon that is related to memorizing the training data and that is well known in statistical learning is called \textit{overfitting}. It amounts to the trained function fitting the available  data (too) well, while not generalizing to unseen data, as illustrated in the bottom left panel of \Cref{fig: underfitting overfitting}. The classical viewpoint in statistics is that when the function has far more parameters than there are data points (as is common with deep neural networks) and if the training time is too large, overfitting might happen, as illustrated in \Cref{fig:training}. An indication of overfitting can be that the generalization error is strongly growing while the empirical risk is driven almost to zero. To prevent this, an alternative to increasing the number of training steps, $t$, while the training data remains the same, is early stopping. It has been shown (e.g. \cite[Cor.~7]{yang2021generalization}) that the empirical distribution can be close to the truth (in which case the ANN generalizes well), if the training is stopped after a sufficiently long, but not too long training phase. Figure \ref{fig:training} shows the typical shape of the discrepancy between the trained network and the truth.\par\bigskip

\begin{figure}[H]
\centering
\includegraphics[width=0.45\linewidth]{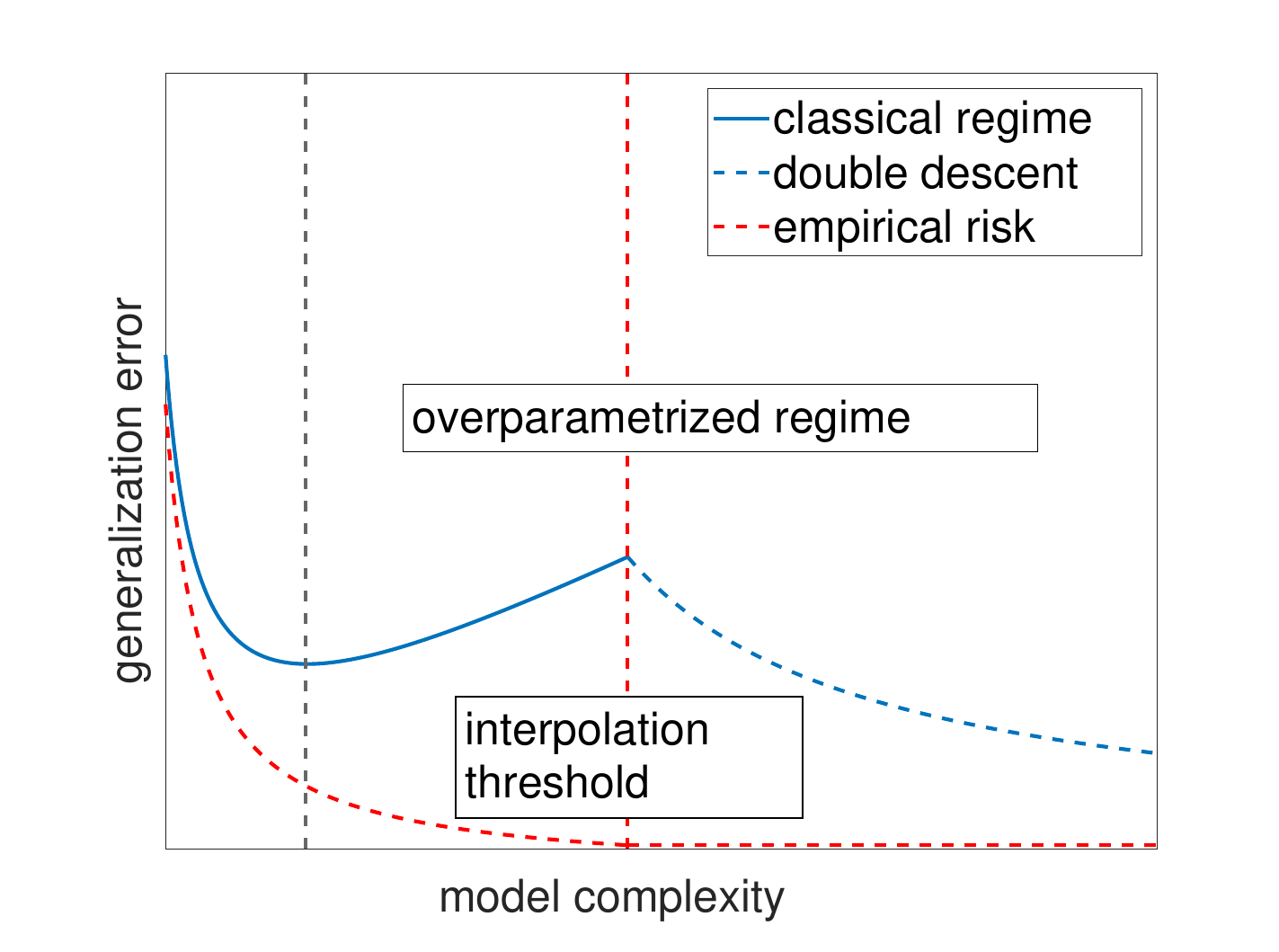}
\caption{Risk curve with benign overfitting: highly overparametrized ANNs often exhibit the double descent phenomenom when the number of parameters exceeds the number of data points. The leftmost vertical dashed line shows the optimal model complexity (for given observation data), beyond which the model is considered overparametrized. The rightmost vertical dashed line marks the interpolation threshold at which the model can exactly fit all data points.}
\label{fig:doubledescent}
\end{figure}

However, it turns out that there are also cases of benign overfitting, in which an ANN shows remarkable generalization properties, even though it is essentially fitting the noise in the training data. The phenomenon of benign overfitting, also known by the name of \emph{double descent}, describes the empirical observation that the generalization error, as measured by the true risk,  decreases again as the number of parameters is increased -- despite severe overfitting (see Figure \ref{fig:doubledescent}). Note that there is not contradiction between the double descent phenomenon and the traditional U-shaped risk curve shown in Figure \ref{fig:training} as they hold under different circumstances and the double descent requires pushing the number of parameters beyond a certain (fairly large) threshold.   

It has been conjectured that this phenomenon is related to a certain low rank property of the data covariance; nevertheless a detailed theoretical understanding of the double descent curve for a finite amount of training data is still lacking as the available approximation results cannot be applied in situations in which the number of parameters is much higher than the number of data points. Interestingly, double descent has also been observed for linear regression problems or kernel methods, e.g. \citep{bartlett2020overfitting,mei2021double}. Thus it does not seem to be a unique feature of ANNs; whether or not it is a more typical phenomenon for ANNs is an open question though \citep{belkin2019tradeoff}; see also \citet{opper1990perceptron} for an early reference in which the double descent feature of ANNs has been first described (for some models even multiple descent curves are conjectured \citep{chen2020multiple, liang2020multiple}). 

\subsection{Curse of dimensionality}

An important aspect of function approximation (and therefore related to perspective 2 stated in \Cref{sec: probabilistic modelling}) is the question of how complicated the function $f_\theta$ or, equivalently, how rich the function class $\mathcal{F}$ needs to be. This becomes particularly interesting if the state space is high-dimensional and a notorious challenge is known as the \emph{curse of dimensionality}. It describes the phenomenon that approximating a target function $f^B$ or the corresponding probability distribution $Q^B=Q(f^B)$ when $\mathcal{X}$ is high-dimensional (i.e. when the number of degrees of freedom is large) requires a huge amount of training data to determine a regression function $f_\theta$ that is able to approximate the target. 
As a rule of thumb, approximating a function $f^B$ on $\mathcal{X}=\R^d$ or the associated probability measure $Q^B$ with an accuracy of $\epsilon$ needs about
\[N=\epsilon^{-\Omega(d)}\] 
sample points in order to determine roughly the same number of a priori unknown parameters $\theta$, thereby admitting an exponential dependence on the dimension.\footnote{Here we use the Landau notation $\Omega(d)$ to denote a function of $d$ that asymptotically grows like $\alpha\cdot d$ for some constant $\alpha>0$; often $\alpha=1,2$.}  
It is easy to see that the number of parameters needed and the size of the training set become astronomical for real-world tasks. 
As an example, consider the classification of handwritten digits. The MNIST database (Modified National Institute of Standards and Technology database) contains a dataset of about 60\,000 handwritten digits that are stored in digital form as $28\times 28$ pixel greyscale images \citep{lecun1998mnist}. If we store only the greyscale values for every image as a vector, then, the dimension of every such vector will be $28^2=784$. By today's standards, this is considered a small system, yet it is easy to see that training a network with about $10^{784}$ parameters and roughly the same number of training data points is simply not feasible, especially as the training set contains less than $10^5$ data points. 

In practice, the number of ANN parameters and the number of data points needed to train a network can be much smaller. In some cases, this inherent complexity reduction present in deep learning can be mathematically understood. Clearly, when the target function is very smooth, symmetric or concentrated, it is possible to approximate it with a parametric function having a smaller number of parameters. The class of functions that can be approximated by an ANN without an exponentially large number of parameters, however, is considerably larger; for example, Barron-regular functions that form a fairly large class of relevant functions can be approximated by ANNs in arbitrary dimension with a number of parameters that is independent of the dimension \citep[Thm.~3]{barron1993approximation}; there are, moreover, results that show that it is possible to  express \emph{any} labelling of $N$ data points in $\R^d$ by an ANN two layers and in total $p=2N+d$ parameters \citep[Thm.~1]{zhang2021understanding}; cf.~\citep{devore2021approx}. In general, however, the quite remarkable expressivity of deep neural networks with a relatively small number of parameters and even smaller training sets is still not well understood \citep[Sec.~4]{berner2021modern}.\footnote{Here, `relatively small' must be understood with respect to the dimension of the training data set. An ANN that was successfully trained on MNIST data may still have several hundred millions or even billions of parameters; nevertheless, the number of parameters is small compared to what one would expect from an approximation theory perspective, namely $10^{784}$. However, it is large compared to the minimum number of parameters needed to fit the the data, which in our example would be $p=2\cdot 60\,000+784=120\,784$, hence an ANN with good generalization capacities is typically severely overfitting, especially if we keep in mind that the effective dimension of the MNIST images that contains about 90\% black pixels is considerably smaller} 

\subsection{Stochastic optimization as implicit regularization}\label{sec:regularize}

Let us finally discuss an aspect related to the optimization of ANNs (cf. perspective 3 in \Cref{sec: probabilistic modelling}) that interestingly offers a connection to function approximation as well. Here, the typical situation is that no a priori information whatsoever about the function class to which $f^B$ belongs is available. A conventional way then to control the number of parameters and to prevent overfitting is to add a regularization term to the loss function that forces the majority of the parameters to be zero or close to zero and hence effectively reduces the number of parameters \citep{tibshirani1996lasso}. Even though regularization can improve the generalization capabilities, it has been found to be neither necessary nor sufficient for controlling the generalization error \citep{geron2017handson}.
Instead, surprisingly, there is (in some situations proveable) evidence that SGD introduces an implicit regularization to the empirical risk minimization that is not present in the exact (i.e.~deterministic) gradient descent \citep{ali2020implicit,roberts2021sgd}. A possible explanation of this effect is that the inexact gradient evaluation of SGD introduces some noise that prevents the minimization algorithm from getting stuck in a bad local minimum. It has been observed that the effect is more pronounced when the variance of the gradient approximation is larger,  in other words: when the approximation has a larger sampling error \citep{keskar2016large}. A popular, though controversial explanation is that noisier SGD tends to favor wider or flatter local minima of the loss landscape that are conventionally associated with better generalization capabilities of the trained ANN \citep{dinh2017sharp,hochreiter1997flat}. How to unambigously characterize the `flatness' of local minima with regard to their generalization capacities, however, is still an open question. Furthermore, it should be noted that too much variance in the gradient estimation is not favorable either, as it might lead to slower training convergence, and it will be interesting to investigate how to account for this tradeoff; cf.~\citep{bottou2018optimization, richter2020vargrad}.

\begin{example}
To illustrate the implicit regularization of an overfitted ANN by SGD, we consider the true function $f(x) = \sin(2 \pi x)$ and create $N=100$ noisy data points according to $y_n = f(x_n) + 0.15 \,\eta_n$, where $x_n$ is uniformly distributed in the interval $[0,2\pi]$ (symbolically: $x_n\sim \mathcal{U}([0, 2])$) and $\eta_n \sim \mathcal{N}(0, 1)$. We choose a fully connected NN with three hidden layers (i.e. $L=4$), each with $10$ neurons. 

Once we train with gradient descent and once we randomly choose a batch of size $N_b = 10$ in each gradient step. In \Cref{fig: GD vs. SGD} we can see that running gradient descent on the noisy data leads to overfitting, whereas stochastic gradient descent seems to have some implicit regularizing effect.

\begin{figure}[H]
\centering
\includegraphics[width=0.8\linewidth]{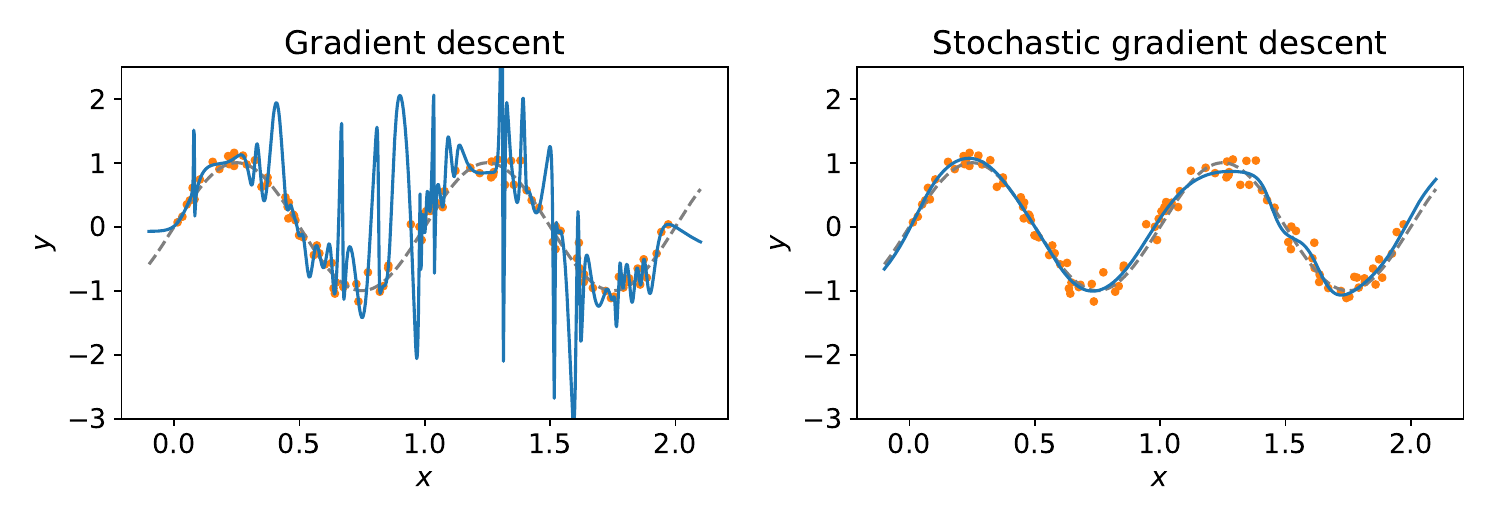}
\caption{We consider a fully connected neural network (blue) that has been trained on $N=100$ noisy data points (orange), once by gradient descent and once by stochastic gradient descent, and compare it to the ground truth function (grey).}
\label{fig: GD vs. SGD}
\end{figure}
\end{example}

We have provided a potpourri of aspects related to the three perspectives \textit{generalization}, \textit{function approximation} and \textit{optimization}, demonstrating subtleties of deep learning that have partly been understood with the help of rigorous mathematical analysis, while still leaving many open questions for future research. In the following, let us move towards perspectives 4 and 5 that we have stated in \Cref{sec: probabilistic modelling}. In particular, the following chapter will argue that relying on classical statistical learning theory might not be sufficient in certain practical applications and additional effort and analysis are needed in order to make deep learning more robust.

\section{Sensitivity and (non-)robustness of neural networks}
\label{sec: adversarial attacks}
So far we have measured the performance of prediction models in an `average' sense. In particular we have stated the goal of a machine learning algorithm to minimize the expected loss 
\begin{equation}
    \mathcal{L}(f) = \E\!\left[\ell(f(X), Y) \right],
\end{equation}
where the deviations between predictions and ground truth data are averaged over the (unknown) probability distribution $\P$. Statements from statistical learning theory therefore usually hold the implicit assumption that future data comes from the same distribution and is hence similar to that encountered during training (cf. \Cref{sec: probabilistic modelling}). This perspective might often be valid in practice, but falls short of atypical data in the sense of having a small likelihood, which makes such an occurence a rare event or a \emph{large deviation}. 
Especially in safety-critical applications one might not be satisfied with average-case guarantees, but rather strives for worst-case analyses or at least for an indication of the certainty of a prediction (which we will come back to in the next section). Moreover, it is known that models like neural networks are particularly sensitive with respect to the input data, implying that very small, barely detectable changes of the data can drastically change the output of a prediction model -- a phenomenon that is not respected by an analysis  based on expected losses.

\subsection{Adversarial attacks}\label{sec:imageattack}

An extreme illustration of the sensitivity of neural networks can be noted in \textit{adversarial attacks}, where input data is manipulated in order to mislead the algorithm.\footnote{This desire to mislead the algorithm is in accordance with Popper's dictum that we are essentially learning from our mistakes. As \citet[p.~324]{popper1984suche}) mentions in the seminal speech \emph{Duldsamkeit und intellektuelle Verantwortlichkeit} on the occasion of receiving the \emph{Dr. Leopold Lucas Price of the University of Tübingen} on the 26th May 1981: ``[\ldots] es ist die spezifische Aufgabe des Wissenschaftlers, nach solchen Fehlern zu suchen. Die Feststellung, daß eine gut bewährte Theorie oder ein viel verwendetes praktisches Verfahren fehlerhaft ist, kann eine wichtige Entdeckung sein.''} Here the idea is to add very small and therefore barely noticeable perturbations to the data in such a way that a previously trained prediction model then provides very different outputs. In a classification problem this could for instance result in suggesting different classes for almost identical input data. It has gained particular attention in image classification, where slightly changed images can be misclassified, even though they appear identical to the original image for the human eye, e.g.~\citep{brown2017adversarial,goodfellow2014explaining,kurakin2016adversarial}.

Adversarial attacks can be constructed in many different ways, but the general idea is usually the same. We discuss the example of a trained classifier: given a data point $x\in \R^d$ and a trained neural network $f_\theta$, we add some minor change $\delta \in \R^d$ to the input data $x$, such that $f_\theta(x + \delta)$ predicts a wrong class. One can differentiate in targeted and untargeted adversarial attacks, where either the wrong class is specified or the misclassification to any arbitrary (wrong) class is aimed at. We focus on the former strategy as it turns out to be more powerful. Since the perturbation is supposed to be small (e.g. for the human eye), it is natural to minimize the perturbation $\delta$ in some suitable norm (e.g. the Euclidean norm or the maximum norm) while constraining the classifier to assign a wrong label  $\widetilde{y} \neq y$ to the perturbed data and imposing an additional box constraint. In the relevant literature (e.g.~\citet{szegedy2014intriguing}), an adversarial attack is constructed as the solution to the following optimization problem: 
\begin{equation}\label{eq:aabox}
    \text{minimize } \| \delta \| \qquad \text{subject to} \qquad f_\theta(x + \delta) = \widetilde{y}\quad\text{and} \quad x+\delta\in[0,1]^d\,.
\end{equation}
Note that we have the hidden constraint $f_\theta(x)=y$, where $y\neq \widetilde{y}$ and the input variables have been scaled such that $x\in[0,1]^d$. 
In order to have an implementable version of this procedure, one usually considers a relaxation of (\ref{eq:aabox}) that can be solved with (stochastic) gradient descent-like methods in $\delta$; see e.g.~\citep{carlini2017towards}. 

Roughly speaking, generating an adversarial attack amounts to doing a (stochastic) gradient descent in the data rather than the parameters, with the aim of finding the closest possible input $\widetilde{x}$ to $x$ that gets wrongly classified and to analyze what went wrong.\footnote{Again, quoting  \citet[p.~325]{popper1984suche}: ``Wir müssen daher dauernd nach unseren Fehlern Ausschau halten. Wenn wir sie finden, müssen wir sie uns einprägen; sie nach allen Seiten analysieren, um ihnen auf den Grund zu gehen.''.}

\begin{example}[Adversarial attack to image classification]
Let us provide an example of an adversarial attack in image classification. For this we use the Inception-v3 model from \citep{szegedy2016rethinking}, which is pretrained on $1000$ fixed classes. For the image in the left panel of \Cref{fig: adversarial attack} a class is predicted that seems close to what is in fact displayed. We then compute a small perturbation $\delta$, displayed in the central panel, with the goal of getting a different classification result. The right panel displays the perturbed image $x+\delta$, which notably looks indistinguishable from the original image, yet gets classified wrongly with the same Inception-v3 model. The displayed probabilities are the so-called softmax outputs of the neural network for the predicted classes and they represent some sort of certainty scores.

\begin{figure}[H]
\centering
\includegraphics[width=0.8\linewidth]{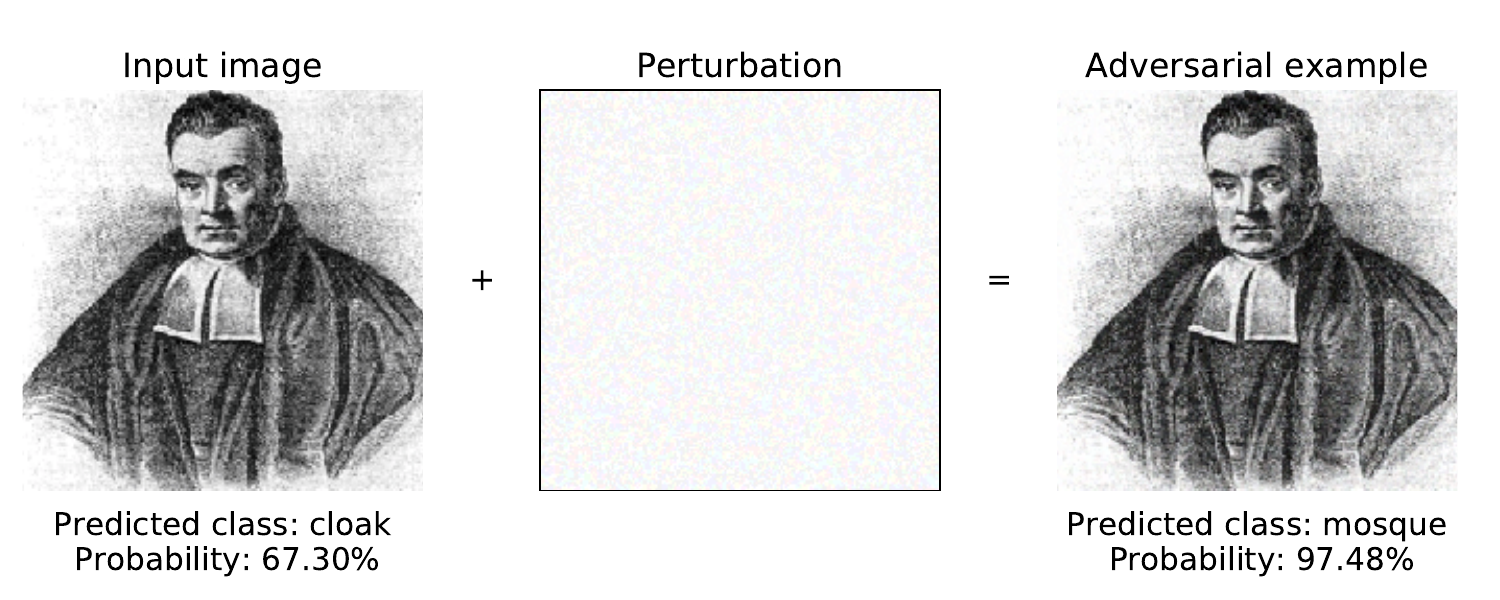}
\caption{The original image of Thomas Bayes in the left panel gets reasonably classified (``cloak''), whereas the right picture is the result of an adversarial attack and therefore gets misclassified (as ``mosque'').}
\label{fig: adversarial attack}
\end{figure}
\end{example}

\subsection{Including worst-case scenarios and marginal cases}
\label{sec: worst-case scenarios}
Adversarial attacks demonstrate that neural networks might not be robust with respect to unexpected input data and the next question naturally is how this issue can be addressed. In fact, multiple defense strategies have been developed in recent years in order to counteract attacks, while it is noted  that a valid evaluation of defenses against adversarial examples turns out to be difficult, since one can often find additional attack strategies afterwards that have not been considered in the evaluation \citep{carlini2019evaluating}. One obvious idea for making neural networks more robust is to integrate adversarial attacks into the training process, e.g. by considering the minimization  
\begin{equation}
\min_\theta \E\!\left[ \max_{\delta \in \Delta}\ell(f_{\theta}(X + \delta), Y)\right] ,
\end{equation}
where $\Delta = \{ \delta : \| \delta \| \le \varepsilon\}$ is some specified perturbation range  \citep{madry2018towards,wang2019convergence}.  Depending on the application, however, convergence of this min-max problem can be cumbersome. At present, the study of adversarial attacks is a prominent research topic with many questions still open (e.g. the role of regularization   \citep{roth2019adversarial}), and it has already become apparent that principles that hold for the average case scenario might not be valid in worst-case settings anymore; cf.~\citep[Sec.~4]{ilyas2019adversarial}. To give an example, there is empirical evidence that overfitting might be more harmful when adversarial attacks are present, in that overparametrized deep NNs that are robust against adversarial attacks may not exhibit the typical double descent phenomenon when the training is continued beyond the interpolation threshold (cf.~Figure \ref{fig:doubledescent}); instead they show a slight increase of the generalization risk when validated against test data, i.e. their test performance degrades, which is at odds with the observations made for standard deep learning algorithms based on empirical risk minimization  \citep{rice2020overfitting}.

Another way to address adversarial attacks is to incorporate uncertainty estimates in the models and hope that those then indicate whether perturbed (or out of sample) data occurs. Note that the question as to whether some new data is considered typical or not (i.e. an outlier or a marginal case) depends on the parameters of the trained neural network which are random, in that they depend on the random training data. As a principled way of uncertainty quantification we will introduce the Bayesian perspective and Bayesian Neural Networks (BNNs) in the next section. We claim that these can be viewed as a more robust deep learning paradigm, which promises fruitful advances, backed up by some already existing theoretical results and empirical evidence. In relation to adversarial attacks, there have been multiple indications of attack identifications \citep{rawat2017adversarial} and improved defenses \citep{feinman2017detecting,liu2018adv,zimmermann2019comment} when relying on BNNs. In fact, there is clear evidence of increasing prediction uncertainty with growing attack strength, indicating the usefulness of the provided uncertainty scores. On the theoretical side, it can be shown that in the (large data and overparametrized) limit BNN posteriors are robust against gradient-based adversarial attacks \citep{carbone2020robustness}.

\section{The Bayesian perspective}
\label{sec: Bayesian perspective}

In the previous chapter we demonstrated and discussed the potential non-robustness of neural networks related, for example, to small changes of input data by adversarial attacks. A connected inherent problem is that neural networks usually \textit{don't know when they don't know}, meaning that there is no reliable quantification of prediction uncertainty.\footnote{Freely adapted from the infamous 2002 speech of the former U.S.~Secretary of Defense, Donald Rumsfeld: ``We [\ldots] know there are known unknowns; that is to say we know there are some things we do not know. But there are also unknown unknowns—the ones we don't know we don't know.''} In this chapter we will argue that the Bayesian perspective is well suited as a principled framework for uncertainty quantification, thus holding the promise of making machine learning models more robust; see \citep{neal1995bayesian} for an overview.

We have argued that classical machine learning algorithms often act as black boxes, i.e.~without making predictions interpretable and without indicating any level of confidence. Given that all models are learnt from a finite amount of data, this seems rather naive and it is in fact desirable that algorithms should be able to indicate a degree of uncertainty whenever not `enough' data have been present during training (keeping in mind, however, that this endeavor still leaves certain aspects of interpretability such as post-hoc explanations \citep[Sec.~3]{du2020posthoc}) open. To this end, the Bayesian credo is the following: we start with some beforehand (\textit{a priori}) given uncertainty of the prediction model $f$. In the next step, when the model is trained on data, this uncertainty will get `updated' such that predictions `close' to already seen data points become more certain. In mathematical terms, the idea is to assume a prior probability distribution $p(\theta)$ over the parameter vector $\theta$ of the prediction model rather than a fixed value as in the classical case. We then condition this distribution on the fact that we have seen a training data set $\mathcal{D} = (x_n, y_n)_{n=1}^N$. 

The computation of conditional probabilities is governed by Bayes' theorem, yielding the \textit{posterior probability} $p(\theta | \mathcal{D})$, namely by
\begin{equation}
\label{eq: Bayes thm parameters}
    p(\theta | \mathcal{D}) = \frac{p(\mathcal{D} | \theta)p(\theta)}{p(\mathcal{D})},
\end{equation}
where $p(\mathcal{D} | \theta)$ is the likelihood of seeing data $\mathcal{D}$ given the parameter vector $\theta$ and $p(\mathcal{D}) = \int_{\R^p} p(\mathcal{D}, \theta) \mathrm d\theta$ is the normalizing constant, sometimes called evidence, assuring that $p(\theta | \mathcal{D})$ is indeed a probability density. The posterior probability can be interpreted as an updated distribution over the parameters given the data $\mathcal{D}$. Assuming that we can sample from it, we can then make subsequent predictions on unseen data $x$ by
\begin{equation}
\label{eq: prediction from BNN}
    f(x) = \int_{\R^p} f_\theta(x) p(\theta | \mathcal{D}) \mathrm d \theta
    \approx \frac{1}{K} \sum_{k=1}^K f_{\theta^{(k)}}(x)
\end{equation}  
where $\theta^{(1)},\ldots,\theta^{(K)}$ are i.i.d.~samples drawn from the Bayesian posterior  $p(\theta | \mathcal{D})$, i.e. we average predictions of multiple neural networks, each of which having parameters drawn from the posterior distribution.

\begin{figure}[H]
\centering
\includegraphics[width=1.0\linewidth]{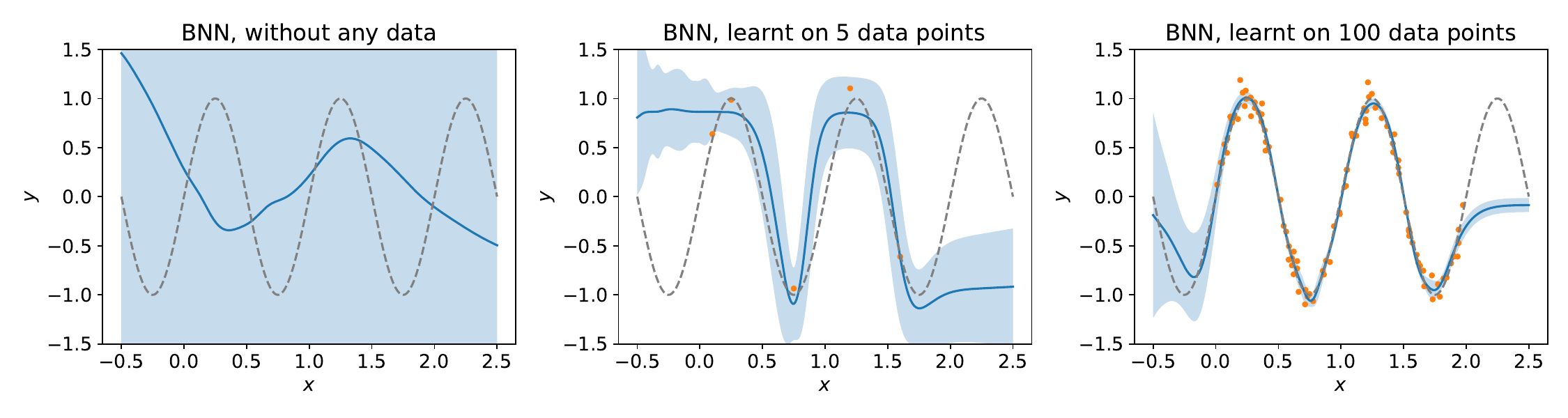}
\caption{We display the evaluation of a BNN by showing its mean prediction function (in dark blue) and a set of two standard deviations from it (in light blue), compared to the ground truth (in gray). Our BNN is either untrained (left) or has seen $N=5$ (in the central panel) or $N=100$ data points (in right panel) during training.}
\label{fig: BNN on data from sine function}
\end{figure}

\begin{example}[BNN based on different amounts of data]
Let us say we want to learn the function $f(x) = \sin(2\pi x)$ and have a certain amount of training data $\mathcal{D} = (x_n, y_n)_{n=1}^N$ available, where the label is given by $y_n = f(x_n) + \eta_n$, with noise $\eta_n \sim \mathcal{N}(0, 0.01)$. Intuitively, the fitted neural network should be closer to the true function as well as more certain in its predictions the more data points are available. We consider a BNN trained with a mean field variational inference attempt and a Gaussian prior on the parameters (see next section for details). In \Cref{fig: BNN on data from sine function} we display the mean prediction function as defined in \eqref{eq: prediction from BNN} as well as a confidence set defined by two standard deviations on the sample predictions. In the left panel we display the evaluation before training, i.e. without relying on any available data points, and note that the average prediction function is rather arbitrary and the uncertainty rather high, as expected. The central panel repeats the same evaluation, where now the BNN is trained on $N = 5$ data points. We can see an improved prediction function and a decreased uncertainty. Finally, the right panel displays a BNN trained on $N = 100$ data points, where now the mean prediction function is quite close to the true function and the uncertainty almost vanishes close to the data points, yet remains large wherever no training data was available. The BNN is therefore able to output reasonable uncertainty scores, depending on which data was available during training.
\end{example}

\subsection{Bayesian neural networks in practice}
\label{sec: BNN in practice}

Even though simple at first glance, the Bayes formula \eqref{eq: Bayes thm parameters} is non-trivial from a computational point of view and can in almost all cases not be computed analytically. The challenging term is $p(\mathcal{D})$, for which, given the nested structure of neural networks, the integral has to be approximated numerically. Classical numerical integration, however, is infeasible too, due to the high dimension of the parameter vector $\theta$. We therefore have to resort to alternative attempts that aim to approximate the posterior distribution $p(\theta | \mathcal{D})$. 

An asymptotically exact method for creating samples from any (suitable) probability distribution is called Hamiltonian Monte Carlo (also: Hybrid Monte Carlo), which is based on ideas from Statistical  Physics and the observation that certain dynamical systems admit an equilibrium state that can be identified with the posterior probability that we seek to compute \citep{neal2011mcmc}. For our purposes this attempt seems to be a method of choice when aiming for high approximation quality; however, it does not scale well to high dimensions and is therefore practically useless for state-of-the-art neural networks. A similar idea is to exploit the so-called Langevin dynamics in combination with subsampling of the data points \citep{welling2011bayesian,zhang2020amagold}. This method scales much better, but it is biased since the data subsampling perturbs the stationary distribution. A quite different attempt is called \textit{dropout}, which builds the posterior approximation into the neural network architecture and implicitly trains
multiple models at the same time \citep{gal2016dropout}. Finally, another popular method is based on variational inference, where the true posterior is approximated within a family of simpler probability densities, e.g. multidimensional Gaussians with diagonal covariance matrices \citep{blundell2015weight}. Depending on the approximation class, this method scales well, but approximation quality cannot be guaranteed.

Each of the methods mentioned above has advantages and disadvantages and many questions are still open. As a general remark, there is indeed repeated evidence that, ignoring the approximation challenges for the moment, the Bayesian framework works well in principle for quantifying the prediction uncertainties of neural networks. Additionally, there are indications, based on empirical studies  \citep{izmailov2021bayesian}, that the overall model performance might be improved when relying on predictions from BNNs in contrast to deterministic ANNs. On the other hand, many of the approximation steps that lead to a BNN are not well understood theoretically, and one can demonstrate empirically that they often lead to posterior approximations that are not accurate, e.g. \citep{foong2019expressiveness}. Some of those failures seem to be due to systematic simplifications in the approximating family \citep{yao2019quality}. This phenomenon gets more severe, while at the same time harder to spot, when the neural networks are large, i.e. when the parameter vector is very high-dimensional. An accepted opinion seems to be that whenever BNNs do not work well, then it is not the Bayesian paradigm that is to blame, but rather the inability to approximate it well \citep{gal2018sufficient}. At the same time, however, there are works such as \citep{farquhar2020liberty} that claim that for certain neural network architectures simplified approximation structures get better the bigger (and in particular the deeper) the model is.
    
\subsection{Challenges and prospects for Bayesian Neural Networks}
\label{sec: challenges for BNNs}

The previous section sought to argue that there is great potential in using BNNs in practice; however, many questions, both from a theoretical and practical point of view, are still open. 
 A natural formulation of BNNs can be based on free energy as a loss function that has been discussed in connection with a formal account of curiosity and insight in terms of Bayesian inference (see \citet{firston2017curiosity}):  while the expected loss or risk in deep learning can be thought of as an energy that describes the goodness-of-fit of a trained ANN to some given data (where minimum energy amounts to an optimal fit), the free energy contains an additional entropy term that accounts for the inherent parameter uncertainty and has the effect of smoothing the energy landscape. The result is a trade-off between an accurate fit, which  bears the risk of overfitting, and reduced model complexity (i.e.~Occam's razor). From the perspective of statistical inference, e.g.~\citep{freeEnergyLN}, the free energy has the property that its unique minimizer in the space of probability measures is the sought Bayesian posterior \citep{hartmann2017vari}. Selecting a BNN by free energy minimization therefore generates a model that, \emph{on average}, provides the best explanation for the data at hand, and thus it can be thought of as making an inference to the best explanation in the sense of \citet{harman1965}; cf. also \citep{mcauliffe2015}.

Evidently, the biggest challenge seems to be a computational one: how can we approximate posterior distributions of large neural networks both well and efficiently? 
But even if the minimizer or the Bayesian posterior can be approximated, the evaluation of posterior accuracy (e.g. from the shape of the free energy in the neighborhood of the minimizer) is still difficult and one usually does not have clear guarantees. Furthermore, neural networks keep getting larger and more efficient methods that can cope with ever higher dimensionality are needed. 
Regarding the benefits of BNNs, there is an open debate on how much performance gains they actually bring in practice; cf.~\citep{wenzel2020good}. Uncertainty quantification, on the other hand, is valuable enough to continue the Bayesian endeavor, eventually allowing for safety-critical applications or potentially improving active and continual learning.

\section{Concluding remarks}
\label{sec: conclusion}

The recent progress in artificial intelligence is undeniable and the related improvements in various applications are impressive. This article, however, provides only a snapshot of the current state of deep learning, and we have demonstrated that many phenomena that are intimately connected are still not well understood from a theoretical point of view. We have further argued that this lack of understanding not only slows down further systematic developments of practical algorithms, but also bears risks that become in particular apparent in safety-critical applications. While inspecting deep learning from the mathematical angle, we have highlighted five perspectives that allow for a more systematic treatment, offering already some novel explanations of striking observations and bringing up valuable questions for future research (cf. \Cref{sec: probabilistic modelling}). 

We have in particular emphasized the influence of the numerical methods on the performance of a trained neural network and touched upon the aspect of numerical stability, motivated by the observation that neural networks are often not robust (e.g. with respect to unexpected input data or adversarial attacks) and do not hold any reliable measure for uncertainty quantification. As a principled framework that might tackle those issues, we have presented the Bayesian paradigm and in particular Bayesian neural networks, which provide a natural way of quantifying epistemic uncertainties. In theory, BNNs promise to overcome certain robustness issues and many empirical observations are in line with this hope; however, they also bring additional computational challenges, connected mainly to the sampling of high dimensional probability distributions. The existing methods addressing this issue are neither sufficiently understood theoretically nor produce good enough (scalable) results in practice such that a persistent usage in applications is often infeasible. We believe that the theoretical properties of BNNs (or ANNs in general) cannot be fully understood without understanding the numerical algorithms used for training and optimisation. Future research should therefore aim at improving these numerical methods in connection with rigorous approximation guarantees. 

Moreover, this article argued that many of the engineering-style improvements and anecdotes related to deep learning need systematic mathematical analyses in order foster a solid basis for artificial intelligence\footnote{This view addresses the skeptical challenge of Ali Rahimi who gave a presentation at NeurIPS Conference in 2017 with the title ``Machine learning has become alchemy''. According to Rahimi, machine learning and alchemy both work to a certain degree, but the lack of theoretical understanding and interpretability of machine learning models is major cause for concern.}. Rigorous mathematical inspection has already led to notable achievements in recent years, and in addition to an ever enhancing handcrafting of neural network architectures, the continuation of this theoretical research will be the basis for further substantial progress in machine learning. We therefore conclude with a quote from Vladimir \citet[p.~X]{vapnik1999nature}, one of the founding fathers of modern machine learning: ``I heard reiteration of the following claim: Complex theories do not work, simple algorithms do. [...] I would like to demonstrate that in this area of science a good old principle is valid: Nothing is more practical than a good theory.'' 

\par\bigskip

\textbf{Acknowledgements.} This research has been partially funded by Deutsche Forschungsgemeinschaft (DFG) through the grant CRC 1114 ‘Scaling Cascades in Complex Systems’ (project A05, project number 235221301) as well as by the Energy Innovation Center (project numbers 85056897 and 03SF0693A) with funds from the Structural Development Act (Strukturstärkungsgesetz) for coal-mining regions.

\newpage

\appendix

\section{Training of artificial neural networks}
\label{sec:SGD}

Let $\mathcal{F}$ be the set of neural networks $f_\theta=\Phi_\sigma$ of a certain predefined topology (i.e.~with a given number of concatenated activation functions, interconnection patterns, etc.) that we want to train. Suppose we have $N$ data points $(x_1, y_1),\ldots,(x_N,y_N)$ where, for simplicity, we assume that $y_n=f(x_n)$ is deterministic. For example, we may think of every $x_n$ having a unique label $y_n=\pm 1$. 
Training an ANN amounts to solving the regression problem 
\[
f_\theta(x_n)\approx y_n
\] 
for all $n=1,\ldots,N$. Specifically, we seek $\theta\in\Theta$ that minimizes the empirical risk (also: loss landscape)
\[
J_N(\theta) = \frac{1}{N}\sum_{n=1}^N \ell(f_\theta(x_n),y_n)
\]
over some potentially high-dimensional parameter set $\Theta$.\footnote{Recall that we call the empirical risk $J_N$ when considered as a function of parameters $\theta$ and $\mathcal{L}_N$ when considered as a function of functions.} There are few cases in which the risk minimization problem has an explicit and unique solution if the number of independent data points is large enough. One such case in which an explicit solution is available is when $f_\theta(x)=\theta^\top x$ is linear and $l(z,y)=|z-y|^2$ is quadratic. This is the classical linear regression problem. 

For ANNs, an explicit solution is neither available nor unique, and an approximation to $\widehat{f}_N\approx f^*$ must be computed by an suitable iterative numerical method. One such numerical method is called gradient descent
\[
\theta_{k+1} = \theta_k - \eta_k\nabla J_N(\theta_k)\,,\quad k=0,1,2,3,\ldots\,,
\]
where $\eta_0,\eta_1,\eta_2$ is a sequence of step sizes, called \emph{learning rate}, that tends to zero asymptotically. For a typical ANN and a typical loss function, the derivative (i.e. the gradient)
\[
\nabla J_N(\theta) = \frac{1}{N}\sum_{n=1}^N \nabla_\theta \ell(f_\theta(x_n),y_n)
\]
with respect to the parameter $\theta$ can be computed by what is called \emph{backpropagation}, essentially relying on the chain rule of differential calculus; see, e.g. \citep{higham2019deep}. Since the number of training points, $N$, is typically very large,  evaluating the gradient that is a sum of $N$ terms is computationally demanding,  therefore the sum over the training data is replaced by a sum over a random, usually small subsample of the training data. This means that, for fixed $\theta$, the derivative $\nabla J_N(\theta)$ is replaced by an approximation $\nabla \widehat{J}_N(\theta)$ that is random; the approximation has no systematic error, i.e. it equals the true derivative on average, but it deviates from the true derivative by a random amount (that may not even be small, but that is zero on average). As a consequence, we can rewrite our gradient descent as follows: 
\begin{equation}\label{SME}
\theta_{k+1} = \theta_k - \eta_k\nabla \widehat{J}_N(\theta_k)  + \zeta_k\,,\quad k=0,1,2,3,\ldots\,,
\end{equation}
where $\zeta_k$ is the random error invoked by substituting $\nabla J_N(\theta)$ with $\widehat{J}_N(\theta)$. Since $\zeta_k$ is unknown as it depends on the true derivative $\nabla J_N(\theta_k)$ at stage $k$ that cannot be easily computed, the noise term in (\ref{SME}) is ignored in the training procedure, which leads to what is called \emph{stochastic gradient descent (SGD)}: 
\begin{equation}\label{SGD}
\theta_{k+1} = \theta_k - \eta_k\nabla \widehat{J}_N(\theta_k)\,,\quad k=0,1,2,3,\ldots\,.
\end{equation}
Since the right hand side in (\ref{SGD}) is random by virtue of the randomly chosen subsample that is used to approximate the true gradient, the outcome of the SGD algorithm after, say, $t$ iterations will always be random. 

As a consequence, training an ANN for given data and for a fixed number of training steps, $t$, multiple times will never produce the same regression function $f_\theta$, but instead a random collection of regression functions. This justifies the idea of a trained neural as a probability distribution $Q(f(t))=Q(f_{\theta(t)})$ rather than unique function $f(t)=f_{\theta(t)}$ that represents its random state after $t$ training steps. 

We should stress that typically, SGD does not converge to the optimal solution (if it converges at all), but rather finds a suboptimal local optimum (if any). From the perspective of mathematical optimization, it is one of the big mysteries of deep learning that despite being only a random and suboptimal solution, the predictions made by the resulting trained network are often suprisingly good \cite[Sec.~1.3]{berner2021modern}. 
In trying to reveal the origin of this phenomenon, SGD has been analyzed using  asymptotic arguments, e.g.~\citep{weinan2017sme,weinan2019sme,mandt2015continuous}. These methods rely on limit theorems, e.g.~\citep{kushner2003stochastic}, to approximate the random noise term in (\ref{SME}), and they are suitable to understand the performance in the large data setting. However, they are unable to adress the case of finite, not to mention sparse training data. Recently, the finite data situation has been analyzed using backward error analysis, and there is empirical evidence that SGD incorporates an implicit regularization which favors shallow minimization paths that leads to broader minima and (hence) to more robust ANNs \citep{barrett2020implicit,smith2021origin,soudry2018implicit}.

\section{Optimal prediction and Bayes classifier}\label{sec:bayesopt}

For prediction tasks, when the ANN is supposed to predict a quantity $y\in\R$ based on an input $x\in\R^d$, the generalization error is typically measured in the sense of the mean square error (MSE), with the quadratic loss
\[
\ell(f(x),y) = (f(x)-y)^2\,.
\]
Let 
\[
\mathrm{sgn}(z) = \begin{cases}
1\,, & z > 0 \\ 0 \,, &  z=0 \\ -1\,, & z<0\,
\end{cases}
\]
be the sign function. Then, for the binary classification tasks, with $y\in\{-1,1\}$ and a classifier $f(x)=\mathrm{sgn}(h(x))$ for some function $h\colon\R^d\to\R$ the quadratic loss reduces to what is called the 0-1 loss (up to a multiplicative constant):
\[
\frac{1}{4}\ell(f(x),y) = \mathbf{1}_{(-\infty,0]}(yh(x)) = \begin{cases}
0\,, & f(x)=y\\ 1\,, & \textrm{else}\,.
\end{cases}
\]
In this case $\mathcal{L}(f) = \P(Y\neq f(X))$ is simply the probability of misclassification. We define the \emph{regression function}
\[
g(x) = \E[Y|X=x]
\]
to be the conditional expectation of $Y$ given the observation $X=x$. Then, using the properties of the conditional expectation, the MSE can be decomposed in a Pythagorean type fashion as
\begin{align*}
    \E[(f(X)-Y)^2] & =\E[(f(X)-g(X) + g(X) - Y)^2] \\
    & = \E[(f(X)-g(X))^2] + 2\E[(f(X)-g(X))(g(X) - Y)]  +  \E[(g(X) - Y)^2]\\
    & = \E[(f(X)-g(X))^2]  +  \E[(g(X) - Y)^2]\,.
\end{align*}
The cross-term disappears since, by the tower property of the conditional expectation,
\begin{align*}
    \E[(f(X)-g(X))(g(X) - Y)] & = \E[\E[(f(X)-g(X))(g(X) - Y)|X]]\\
    & = \E[\E[(f(X)-g(X))g(X)|X]] - \E[\E[(f(X)-g(X))Y|X]]\\
    & = \E[(f(X)-g(X))g(X)] - \E[(f(X)-g(X))\E[Y|X]]\\
    & = 0 \,.
\end{align*}
As a consequence, we have for all functions $f$: 
\[
\mathcal{L}(f) = \E[(f(X)-g(X))^2]  +  \E[(g(X) - Y)^2] \ge  \E[(g(X) - Y)^2]
\]
where equality is attained if and only if $f=g$. The findings can be summarized in the following two statements that hold with probability one:\footnote{If a statement is said to hold \emph{with probability one} or \emph{almost surely}, this means that it is true upon ignoring events of probability zero.}
\begin{itemize}
    \item[(1)] The regression function is the minimizer of the MSE, i.e. we have $g=f^B$, with unique
    \[
    f^B(x) \in \argmin_{f \in \mathcal{M}(\mathcal{X}, \mathcal{Y})} \E[(f(X)-Y)^2]\,. 
    \]
    \item[(2)] The MSE can be decomposed as
    \[
    \mathcal{L}(f) = \E[(f(X)-\E[Y|X])^2]  +  \mathcal{L}^*\,,
    \]
    where the \emph{Bayes risk} $\mathcal{L}^*=\mathcal{L}(f^B)$ measures the variance of $Y$ for given $X=x$ around its \emph{optimal prediction} 
    \[f^B(x)=\E[Y|X=x]\,.\]  
\end{itemize}
The reasoning carries over to the classification task with $Y\in\{-1,1\}$, in which case \[g(x)=\P(Y=1|X=x)-\P(Y=-1|X=x)\] 
and the \emph{optimal classifier} or \emph{Bayes classifier} can be shown to be
\[
f^B(x) = \mathrm{sgn}(g(x)) = \begin{cases}
1\,, & \P(Y=1|X=x) > \P(Y=-1|X=x)\\ -1\,, & \P(Y=1|X=x) < \P(Y=-1|X=x)\,.
\end{cases}
\]

\bibliographystyle{apalike-limitsai}
\bibliography{references}

\end{document}